\pdfoutput=1
\documentclass[11pt]{article}

\usepackage[]{ACL2023}

\usepackage{times}
\usepackage{latexsym}
\usepackage{todonotes}
\usepackage[T1]{fontenc}

\usepackage[utf8]{inputenc}
\usepackage{CJK}
\usepackage{microtype}

\usepackage{inconsolata}

\usepackage{tabularx}
\usepackage{graphicx}
\usepackage{latexsym}
\usepackage{array,ragged2e}
\newcolumntype{P}[1]{>{\centering\arraybackslash}p{#1}}
\newcolumntype{Y}{>{\centering\arraybackslash}X}
\newcolumntype{L}{>{\raggedright\arraybackslash}X}

\title{FEED PETs: Further Experimentation and Expansion on the Disambiguation of Potentially Euphemistic Terms\\
}

\author{Patrick Lee,  {\bf Iyanuoluwa Shode }, {\bf Alain Chirino Trujillo}, {\bf Yuan Zhao}, \\ {\bf Olumide Ebenezer Ojo}, {\bf Diana Cuevas Plancarte},  {\bf Anna Feldman}, {\bf Jing Peng}\\
      Montclair State University\\New Jersey, USA\\
      \texttt{\{leep6,shodei1,chirinotruja1,zhaoy2,ojoo,cuevasplancd1,feldmana,pengj\}}\\
        \texttt{@montclair.edu}}
\begin{document}
\maketitle
\begin{abstract}
Transformers have been shown to work well for the task of English euphemism disambiguation, in which a potentially euphemistic term (PET) is classified as euphemistic or non-euphemistic in a particular context. In this study, we expand on the task in two ways. First, we annotate PETs for vagueness, a linguistic property associated with euphemisms, and find that transformers are generally better at classifying vague PETs, suggesting linguistic differences in the data that impact performance. Second, we present novel euphemism corpora in three different languages: Yoruba, Spanish, and Mandarin Chinese. We perform euphemism disambiguation experiments in each language using multilingual transformer models mBERT and XLM-RoBERTa, establishing preliminary results from which to launch future work. 
\end{abstract}

\section{Introduction}

Detecting and interpreting figurative language is a rapidly growing area in Natural Language Processing (NLP)~\cite{chakrabarty2022flute,10.5555/3298023.3298038}. Unfortunately, little work has been done on euphemism processing. Euphemisms are expressions that soften the message they convey. They are culture-specific and dynamic: they change over time. Therefore, dictionary-based approaches are ineffective \cite{betram1998,holder2002,rawson2003dictionary}. Euphemisms are often ambiguous: their figurative and non-figurative interpretation is often context-dependent; see Table \ref{tab:ambiguity} for examples. Thus, existing work refers to these expressions as potentially euphemistic terms (PETs). State-of-the-art language models such as transformers perform well on many major NLP benchmarks. Recently, an attempt has been made to determine how these models perform in the euphemism disambiguation task \cite{lee2022report}, in which an input text is classified as containing a euphemism or not. The described systems report promising results; however, without further analysis and experimentation, it is unclear what transformers are capturing in order to perform the disambiguation, and the full extent of their ability in other languages.

\begin{table*}[htp]
\begin{small}
    \centering
    \begin{tabular}{l|l}
    Non-euphemistic & Euphemistic\\\hline
Asked to choose \underline{between jobs} and the environment,&This summer, the budding talent agent was\\
a majority -- at least in our warped, &  \underline{between jobs} and free to babysit pretty much \\ 
first-past-the-post system -- will pick jobs. &any time. \\
         \hline
Managers and scientists switch \underline{between jobs} in private & The couple say that they employ some great \\
industry and government in USA in a manner &  baristas and are looking to train more as the \\
 perhaps not yet noticeable in India.& business expands, they emphasise that it \\
& is a job offering a great career and not just\\
&  for students and those \underline{between jobs}. \\
\hline
    \end{tabular}
    \caption{Euphemistic and non-euphemistic interpretations are context-sensitive. \\Ambiguity of \underline{between jobs} (Retrieved from the News on the Web Corpus, October 6, 2021)}
    \label{tab:ambiguity}
    \end{small}
\end{table*}

%however, it is un-clear why large language models (LLMs) behave like that in euphemistic expressions and what features of the euphemisms they capture.%

To address this, the present study describes two experiments to expand upon the euphemism disambiguation task. In the first, we investigate a pragmatic property of euphemisms, vagueness, and use human annotations to distinguish between PETs which are more vague (vague euphemistic terms, or VETs) versus less vague. We then experiment with transformers' abilities to disambiguate examples containing VETs versus non-VETs, and find that performance is generally higher for VETs. While we are unable to ascertain the exact reason for this discrepancy, we analyze the potential implications of the results and propose follow-up studies. In the second experiment, we create novel euphemism corpora for three other languages: Yor\`{u}b\'{a}, (Latin American and Castilian) Spanish, and Mandarin Chinese. Similarly to the English data, examples are obtained using a seed list of PETs, and include both euphemistic and non-euphemistic instances. We run initial experiments using multilingual transformer models mBERT and XLM-RoBERTa, testing their ability to classify them. The results establish preliminary baselines from which to launch future multilingual and cross-lingual work in euphemism processing.

\section{Previous Work}
In the past few years, there has been an interest in the NLP community in computational approaches to euphemisms. \citet{felt2020recognizing} present the first effort to recognize euphemisms and dysphemisms (derogatory terms) using NLP. The authors use the term  \emph{x-phemisms} to refer to both. They used a weakly supervised algorithm for semantic lexicon induction \cite{thelen2002bootstrapping} to generate lists of near-synonym phrases for three sensitive topics (lying, stealing, and firing). The important product of this work is a gold-standard dataset of human x-phemism judgements showing that sentiment connotation and affective polarity are useful for identifying x-phemisms, but not sufficient. 

While the performance of \citet{felt2020recognizing}'s system is relatively low and the range of topics is very narrow, this work inspired other research on euphemism detection.  Thus, \citet{zhu2021self} define two tasks:  1) euphemism detection (based on the input keywords, produce a list of candidate euphemisms) 2) euphemism identification (take the list of candidate euphemisms produced in (1) and output an interpretation). The authors selected sentences matched by a list of keywords, created masked sentences (mask the keywords in the sentences) and applied the masked language model proposed in BERT \cite{devlin2018bert} to filter out generic (uninformative) sentences and then generated expressions to fill in the blank. These expressions are ranked by relevance to the target topic.

\citet{gavidia-etal-2022-cats} present the first corpus of potentially euphemistic terms (PETs) along with example texts from the GloWbE corpus. They also present a subcorpus of texts where these PETs are not being used euphemistically. \citet{gavidia-etal-2022-cats} find that sentiment analysis on the euphemistic texts supports that PETs generally decrease negative and offensive sentiment. They observe cases of disagreement in an annotation task, where humans are asked to label PETs as euphemistic or not in a subset of our corpus text examples. The disagreement is attributed to a variety of potential reasons, including if the PET was a commonly accepted term (CAT). This work is followed by \citet{lee2022searching} who present a linguistically driven proof of concept for finding potentially euphemistic terms, or PETs. Acknowledging that PETs tend to be commonly used expressions for a certain range of sensitive topics, they make use of distributional similarities to select and filter phrase candidates from a sentence and rank them using a set of simple sentiment-based metrics. 

 With regards to the euphemism disambiguation task, in which terms are classified as euphemistic or non-euphemistic, a variety of BERT-based approaches featured in the 3rd Workshop on Figurative Language Processing have shown promising results. \citet{eureka:22} and \citet{kesen2022detecting} both show that supplying the classifier with information about the term itself, such as embeddings and its literal (non-euphemistic) meaning, significantly boost performance, among other enhancements. In a zero-shot experiment, \citet{keh:22} shows that BERT can disambiguate PETs unseen during training (albeit at a lower success rate), suggesting that some form of general knowledge is learned, though it is unclear what.

\section{VET Experiments}

In this section, we discuss the concept of Vague Euphemistic Terms (VETs), and subsequent experiments. The linguistics literature often describes euphemisms as either `more ambiguous' or `vaguer' than the non-euphemistic expressions they substitute \cite{burridge2012euphemism,williamson2002vagueness,egre2011introduction,russell1923vagueness,di2013vagueness}. We understand ambiguity as a countable property, when an expression can have a certain number of senses; whereas vagueness is not countable, a continuum of meaning or theoretically an infinite number of interpretations. However, we note that these qualities are on a "spectrum", and may not be equal for all euphemisms. See below for examples of some euphemisms which may be considered to be VETs, and others, non-VETs:

\begin{table*}[htp]
\begin{small}
    \centering
    \begin{tabular}{l|l}
        Non-euphemistic & Euphemistic \\
        \hline 
        pregnant woman & woman in a certain condition \\
        aged care institution & home, hostel, house, cottage, village, residence \\
        old age & certain age \\
        false statements & alternative facts \\
        war & special military operation/campaign \\
        we have to change and do something we aren't used to & we must reach beyond our fears \\
        being out of work & being in transition \\
        a lack of consistent access to enough food for an active healthy life & food insecurity \\
        prison & correctional facility \\
        blind & visually challenged, visually impaired \\
        \hline
    \end{tabular}
    \caption{Euphemisms are vaguer than the expressions they substitute.}
    \label{tbl:vagueness}
    \end{small}
\end{table*}

\bigskip
\noindent
\emph{VAGUE: The funds will be used to help <neutralize> threats to the operation and ensure our success.} (Counter? Peacefully or violently? Kill? Some other form of removing power?)
\newline
% \centerline{$\downarrow$}
\emph{VAGUE: They were really starting to like each other, but did not know if they were ready to <go all the way> yet.} (Start dating? Have sexual intercourse? Begin or complete some other process?)
\newline
\emph{NONVAGUE: As part of their restructuring, the company will <lay off> part of their workforce by next week.}
\newline
\emph{NONVAGUE: There is always gossip about who <slept with> who on the front page of the magazine.}
\newline

Additionally, \citet{gavidia-etal-2022-cats,lee2022searching} observed that there are different kinds of potentially euphemistic terms (PETs).  One distinction they suggest is `commonly accepted terms' (CATs),  which are so commonly used in a particular domain that they may have less pragmatic purpose (intention to be vague/neutral/indirect/etc.) than other euphemisms. Some examples of PETs which may be CATs are "elderly", "same-sex", and "venereal disease". Humans may disagree on whether these terms are euphemistic in context, since CATs may be viewed as "default terms" rather than a deliberate attempt to be euphemistic. Notably, since many of the PETs under investigation are established expressions, we expect a fair amount to be non-vague; i.e., modern speakers of the language should precisely understand what the term means.

The differences described above may be a factor in computational attempts to work with euphemisms; e.g., some examples may be harder to disambiguate. To investigate this, we assess transformers' performances on examples annotated to be "vague" versus those that are "non-vague". However, defining and determining the relative vagueness of an expression is not a trivial task. Below, we describe our methodology for obtaining vagueness labels, experimental results and follow-up analyses.

\subsection{Methodology}

\subsubsection{Vagueness Labels}
To examine correlations between model performance and vagueness, we first aim to label each PET with a binary label (0 for non-vague, and 1 for vague). Existing computational methods for measuring vagueness are primarily lexically driven, using a dictionary of "vague terms", such as "approximately" or gradable adjectives like "tall" \cite{guelorget:21, lebanoff18}, and do not fit our use case. Thus, we consider human-annotation approaches.  However, in discussions with authors and annotators, we found that there was significant disagreement on what is meant by "vagueness", and how it should be defined for this task. Lacking clear instructions for explicitly annotating vagueness, we opted for an indirect annotation task. In this task, we asked annotators to replace the PET with a more direct paraphrase (if possible), and use similarities in annotators' paraphrases as a proxy for "vagueness". Intuitively, if annotators give dissimilar responses for a particular PET, then this indicates the PET is open to multiple interpretations, and thus a VET.

\begin{table*}[]
\begin{center}
\begin{tabularx}{\textwidth}{|L|c|Y|c|c|} 
\hline
\textbf{Text} & \textbf{Euph Label} & \textbf{Paraphrases} & \textbf{Cos Sim} & \textbf{Vague Label} \\
\hline
The violent Indian <Freedom Fighters> who fought the British were very much this. [...] & 1 & revolutionaries, reformers, anti-government activists, insurrectionists, terrorists, terrorists & 0.53 & 1 \\
\hline
[...] He's <passed away> but he started out as [...] & 1 & dead, died, died, died, died, died & 0.924 & 0 \\
\hline
[...] were electrocuted for <passing on> nuclear information to Soviet Russia [...] [...] & 0 & smuggling, leaking, illegally spreading, giving, passing on, giving away & 0.330 & 1 \\
\hline
At home, I wasn't allowed to watch certain movies until I had reached <a certain age>. [...] & 0 & an old enough age, a certain age, grown mature enough, maturity, adulthood, a certain age & 0.608 & 0 \\
\hline
\end{tabularx}
\end{center}
\caption{Sample of annotation results. The "Paraphrases" column shows the six annotators' responses, and the "Cos Sim" column shows the cosine similarity scores between embeddings of the responses.}
\label{tbl:annotation_examples}
\end{table*}

The way we computed the labels was as follows: 
\begin{enumerate}
    \item We supply annotators with a randomly selected example of each PET from the Euphemism Corpus; if a PET was ambiguous, both a euphemistic and a non-euphemistic example was supplied, resulting in an annotation task of 188 examples. A total of 6 linguistically-trained annotators were recruited. Annotators were then supplied with these instructions:
    
    \emph{"For this task, you will read through text samples and decide how to paraphrase a certain word/phrase in the text. Each row will contain some text in the “text” column containing a particular word/phrase within angle brackets < >. In the “paraphrase” column, please try to replace the word/phrase with a more direct interpretation. If you can’t think of one, then answer with the original word/phrase."}
    \item Sentence-BERT \cite{reimers2019sentencebert} was then used to generate embeddings of the annotators' responses. The cosine similarities between the embeddings were computed for each example and acted as an automatic measure of similarity between responses. See Table \ref{tbl:annotation_examples} for sample responses and the respective cosine similarity scores between them.
    \item While this transformer-based similarity score generally captured semantic similarity well for strong cases of similarity or dissimilarity (e.g., see rows 2 and 3 of Table \ref{tbl:annotation_examples}), we found that there were several "borderline cases" in which the score did not accurately reflect the semantic similarity between responses. For instance, annotators sometimes "over-paraphrased" non-euphemistic examples, providing responses with significant lexical differences (e.g., the non-euphemistic usage of the word "expecting" was paraphrased as "expecting", "anticipating", "foreseeing", etc.), that led to a low cosine score, despite being semantically similar to human judgment. Therefore, based on an examination of such borderline cases, we used the automatic method to assign a label of 0 (non-vague) to examples with a cosine score greater than 0.65, a label of 1 (vague) to examples with a score lower than 0.50, and manually annotated all examples in between. See Table \ref{tbl:annotation_examples} for sample responses, and the label they resulted in. 
    \item Lastly, these labels were generalized to the rest of the dataset under the assumption that euphemistic and non-euphemistic PETs are either vague or non-vague, regardless of context. For example, the euphemistic uses of ``passed away" or ``lay off" are usually non-vague, while ``neutralize" and ``special needs" are usually vague. Table \ref{tbl:num_vague} shows the final distribution of vagueness labels in our dataset when using this procedure.
\end{enumerate}

It should be noted that this is an experimental procedure for approximating human labels of vagueness, in lieu of a more established method. In particular, the generalization that all PETs are vague or not regardless of context is a strong assumption. We leave exploring alternate methods of annotating vagueness for future work.

\begin{table}[htb]
\begin{center}
\begin{tabularx}{\columnwidth}{|c|Y|Y|}
\hline
\textbf{} & \textbf{Vague} & \textbf{Non-Vague}\\
\hline
\textbf{Euphemistic} & 408 & 975 \\
\hline
\textbf{Non-Euphemistic} & 361 & 208 \\
\hline
\end{tabularx}
\end{center}
\caption{Number of vague vs. non-vague examples in the dataset}
\label{tbl:num_vague}
\end{table}

\subsubsection{Data and Model}
\label{section:vague_method}
The euphemism dataset used for the experiments is the one created by \citet{gavidia-etal-2022-cats}. A few modifications were made to several examples we believed to be misclassified. The final dataset contained 1952 examples, of which 1383 are euphemistic and 569 are non-euphemistic, spanning 128 different PETs. 

The model used for all experiments was RoBERTa-base \cite{liu2019roberta}. RoBERTa was fine-tuned on the data using 10 epochs, a learning rate of 1e-5, a batch size of 16; all other hyperparameters were at default values. 

Using the vagueness labels, we run classification tests in which RoBERTa is fine-tuned on both vague and non-vague examples, and then tested on both vague and non-vague examples. Then, we compute performance metrics separately for vague and non-vague examples in the test set for comparison. In the training and test sets, the data was split as evenly as possible across all labels of interest to help eliminate the impact of class imbalance on output metrics. Specifically, samples were randomly selected using the size of the smallest subgroup (vague-euphemistic, nonvague-euphemistic, etc.), and then evenly distributed into training and test sets using an 80-20 split. For example, for the vagueness data shown in Table \ref{tbl:num_vague}, 208 is the size of the smallest subgroup, so 208 examples were randomly selected from all other subgroups for a total of 832 examples (664 train and 168 test); i.e., there were equal amounts of vague-euphemistic, vague-non-euphemistic, etc. examples in both training and test sets. Additionally, the number of unique/ambiguous PETs was approximately the same in all data splits.

\subsection{Experimental Results and Observations}

Table \ref{tbl:vagueness-results} shows the results of the VET experiment, which are metrics (Macro-F1, Precision, and Recall) averaged across 10 different classification runs. As aforementioned, in order to look at the effect of vagueness, we compute metrics for vague and nonvague examples separately; the first row shows the average metrics for the vague test examples in each run, while the second row shows metrics for the non-vague test examples. We observe that the performances are better for the examples marked as vague, rather than non-vague, suggesting that this is a meaningful distinction between examples. 

\begin{table}[!h]
\begin{center}
\begin{tabularx}{\columnwidth}{|c|Y|Y|Y|} 
 \hline
 & \textbf{F1} & \textbf{P} & \textbf{R} \\
 \hline
 Vague & 0.853 & 0.856 & 0.854 \\
 \hline
 Non-vague & 0.793 & 0.805 & 0.795 \\
 \hline
\end{tabularx}
\end{center}
\caption{Results from the vagueness experiments.}
\label{tbl:vagueness-results}
\end{table}

As a consequence of the annotation procedure, the immediate conclusion is that examples containing non-vague PETs (i.e., those which annotators interpreted similarly) are somehow harder to classify, while those containing VETs are easier. However, a concrete explanation of this result remains elusive. An initial hypothesis was that non-vague PETs may be more likely to be PETs which annotators disagreed on in the original dataset \cite{gavidia-etal-2022-cats}, but this was not necessarily the case. 

An error analysis of the most frequently misclassified examples leads us to a potential cause for the comparatively poor performance of the non-vague examples. We noted that a significant proportion of misclassified examples were non-euphemistic examples (which had been consistently misclassified as euphemistic by BERT). PETs in these examples appeared to co-occur with a relatively high number of "sensitive words" - words relating to sensitive topics that people may typically use euphemisms for, such as death, politics, and so on. If certain "sensitive words" are typically associated with euphemistic examples, then examples where this is not the case may mislead the classifier. In an attempt to quantify this, we use the following procedure:

\begin{enumerate}
    \item Using a list of sensitive topics previously used for euphemism work as a starting point \cite{lee2022searching}, we come up with "sensitive word list" comprising of a list of 22 words we believe to represent a range of "sensitive topics". See Appendix \ref{sec:appendix_a} for the full list.
    \item For each example, we go through each word and compute the cosine similarity with the words in our "sensitive word list" using Word2Vec \cite{mikolov2013efficient}. For every comparison that yields a similarity score > 0.5, we add a point to this example's "sensitivity score".
    \item We then isolate the examples which were misclassified 10 or more times in the experiments, and repeat the above. 
\end{enumerate}

 Table \ref{tbl:sensitivty_results} below shows the results of this procedure. Each row shows a particular subgroup (e.g., the first row is for the euphemistic, vague examples), the number of examples in the subgroup, and the mean "sensitivty score" for examples in the subgroup. The last column shows the score normalized by the number of words in each example. 

 \begin{table}[!h]
\begin{center}
\begin{tabularx}{\columnwidth}{|Y|Y|P{1.05cm}|c|Y|Y|} 
 \hline
 \textbf{Euph} & \textbf{Vague} & \textbf{Data-set} & \textbf{Size} & \textbf{Mean Score} & \textbf{Norm Score}\\
 % \textbf{DRs} & \multicolumn{2}{c|}{\textbf{Formants}} & \multicolumn{2}{c|}{\textbf{MFCCs}} & \multicolumn{2}{c|}{\textbf{wav2vec2}}\\
 % \cline{2-7}
 % & ACC & F1 & ACC & F1 & ACC & F1 \\
\hline
 1 & 1 & Full & 408 & \textbf{7.94} & \textbf{0.126} \\

 1 & 0 & Full & 975 & 7.78 & 0.13 \\

 0 & 1 & Full & 361 & 5.59 & 0.094 \\

 0 & 0 & Full & 208 & \textbf{5.56} & \textbf{0.095} \\
\hline
 1 & 1 & Err & 21 & \textbf{3.57} & \textbf{0.056} \\

 1 & 0 & Err & 42 & 4.36 & 0.076 \\

 0 & 1 & Err & 45 & 7.09 & 0.114 \\

 0 & 0 & Err & 35 & \textbf{8.26} & \textbf{0.13} \\
 \hline
\end{tabularx}
\end{center}
\caption{Average sensitivity scores for each subgroup of the full corpus (top 4 rows) versus frequently misclassified examples (bottom 4 rows).}
\label{tbl:sensitivty_results}
\end{table}

The first 4 rows of the dataset show that for the full corpus, sensitivity scores are higher for euphemistic examples than for non-euphemistic, regardless of vagueness. This suggests that, although euphemisms are milder alternatives to sensitive words, they tend to co-occur with other sensitive words in the context. 

In contrast, we observe that this trend is reversed for the frequently misclassified examples (bottom 4 rows). That is, the misclassified euphemistic examples have an unusually low sensitivity score, while non-euphemistic examples have an unusually high score. If BERT has associated sensitive words with the euphemistic label, then it may be "confused" by non-euphemistic examples which have a high occurrence of them, and vice versa. Intuitively, we speculate that this happens more frequently with non-vague examples, because usage of a non-vague PET may correlate with decreased pragmatic intent.

Overall, there appears to be a correlation between the sensitivity score and misclassifed examples. Unfortunately, follow-up experiments involving model interpretability and ablation did not yield concrete results, so we cannot yet claim that BERT is "paying attention" to sensitive words. We leave a more comprehensive investigation to future work. However, the vagueness distinction between PETs indicates that there are linguistic differences between examples that have a concrete impact on model performance. Future work includes investigating other pragmatic features of euphemisms in a similar fashion, such as indirectness or politeness, and in other languages besides English.

\section{Multilingual Experiments}

Euphemism disambiguation thus far has focused on American English. In this section, we describe euphemism disambiguation experiments run on multilingual data. For each of the different languages, native speakers and language experts created a list of PETs, collected example texts for each PET, and annotated each text for whether the PET was being used euphemistically given the context. We then test the classification abilities of multilingual transformer models. The results are intended to show whether multilingual transformer models have the potential to disambiguate euphemisms in languages other than English, and establish preliminary baselines for the task.

\subsection{Datasets}
The data collection and annotation for each language is described below. Note that, while inter-annotator agreement is reported by \cite{gavidia-etal-2022-cats}, we did not have enough annotators to report agreement for each language. However, we assume that the agreement for other languages will be similar to American English, and leave more precise metrics for future work with more annotators.

\begin{table*}[!h]
\begin{small}
    \centering
    \begin{tabular}{p{7cm}|p{7cm}}
    \textbf{Non-euphemistic} & \textbf{Euphemistic}\\\hline
\begin{CJK}{UTF8}{gbsn} 放在手机上看又不\underline{方便}。 \end{CJK} / It is not convenient to read it on the phone. & \begin{CJK}{UTF8}{gbsn} 吃饭时，一人说去\underline{方便}一下。 \end{CJK} / During the meal, a person went to use the bathroom. \\
         \hline
& \\\begin{CJK}{UTF8}{gbsn} \underline{方便}了秦始皇的全国巡游。 \end{CJK} / It made the nation-wide tour convenient for Qin Shi Huang. & \begin{CJK}{UTF8}{gbsn} 于是选择了就近的河边\underline{方便}一下。 \end{CJK} / So he chose to relieve himself right by the river. \\
\hline
    \end{tabular}
    \caption{Examples of euphemistic and non-euphemistic sentences in Mandarin Chinese}
    \label{tab:chinese}
    \end{small}
\end{table*}

\begin{table*}[!h]
\begin{small}
\centering
\begin{tabular}{p{7cm}|p{7cm}}
\textbf{Non-euphemistic} & \textbf{Euphemistic}\\
\hline
Es perfecta para divertirse, \underline{pasar un buen rato} y dejarte llevar por una historia sin m\'{a}s pretensi\'{o}n. / It is perfect to have some fun, have a good time and to let yourself carry by an unpretentious story. & Con el prop\'{o}sito evidente de \underline{pasar un buen rato} con ella. La chica no era muy brillante, pero lo que le faltaba de inteligencia le sobraba en curvas. / With the clear purpose of having a good time with her. The girl was not that brilliant, but her curves overshadowed her intelligence. \\
            \hline

Que los \underline{pocos recursos} disponibles estaban comprometidos para pagar las deudas ocultas. /That the few resources are destined to pay off the hidden debt.  &  
Para que j\'{o}venes de \underline{pocos recursos} logren alcanzar su profesionalizaci\'{o}n en las aulas. /So that poor young students find a way to become professionals at school.\\
\hline
\end{tabular}
\caption{Examples of euphemistic and non-euphemistic sentences in Spanish}
\label{tab:spanish}
\end{small}
\end{table*} 
%Given the rich and diverse nature of the Spanish Language, creating this corpus was an interesting event. First, a list of euphemisms was created. This list contained both, pure euphemistic and non-euphemistic terms. Subsequently, some online corpora were located and examined for suitability. The Real Academia Española (Real Spanish Academy) was the most resourceful platform available in order to develop this task, therefore, this was the one destined to develop for this project. The data inside the corpus comprises the PETs, euphemistic/non-euphemistic terms in context, PET label, example source, bibliography and country of origin. This latter was a key piece of data since focusing on this allowed linguistically to include all ethnic dialect groups and reinforce the richness of the Spanish Language. 

%Even though Spanish is a vastly studied language, when it comes to research related to Natural Language Processing and euphemisms it has a long way to go. We should be able to deeper exploit this rich language. 

% As an example, \textit{"o ti ku"} meaning "\textit{he/she has died}",  can be used in a more subtle manner as "o file se aso bora" or "o pa ipo da" when an individual has passed away. An example of a PET in {Yor\`{u}b\'{a}} used is: "Ki Eledumare gba abo okan gbogi ninu Osere Amuludun arabinrin Aishat Abimbola eniti \textbf{"o file se aso bora"} ni ilu Canada lose to koja". 

\begin{table*}[!h]
    \begin{small}
    \centering
    \begin{tabular}{p{7cm}|p{7cm}}   
        \textbf{Non-euphemistic} & \textbf{Euphemistic} \\
        \hline 
        T\d{\'{a}}iw\`{o}, \d{\'{e}}gb\d{o}n F\`{u}nk\d{\`{e}} r\'{i} \underline{\`{a}lej\`{o}}  r\d{\'{e}} l\'{a}n\`{a} t\'{o} w\'{a} l\'{a}ti \`{i}l\'{u} \`{E}k\'{o}. & Ob\`{i}nrin t\'{i} k\`{o} r\'{i} \underline{\`{a}lej\`{o}} r\d{\'{e}}. 
       \\ Taiwo, Funke's elder sibling saw her visitor who came from Lagos yesterday. & The woman who does not see her menstruation. \\ 
        \hline
        A k\`{o} gb\d{o}d\d{\`{o}} \underline{d\d{\'{a}}k\d{\`{e}}}. & \d{E} sara g\'{i}r\'{i}, b\`{a}b\'{a} ti \underline{d\'{a}k\d{\'{e}}}. \\
        We should not be quiet. & Be brave, father is dead. \\
        \hline
    \end{tabular}  
    \caption{Examples of euphemistic and non-euphemistic sentences in {Yor\`{u}b\'{a}}} 
    \label{tab:yoruba}
    \end{small}
\end{table*}

\subsubsection{Mandarin Chinese}

Euphemisms are widely used in Chinese Mandarin in both formal and informal contexts, and in spoken and written language. It has been a social norm to use euphemisms to express respect and sympathy, and also to avoid certain taboos and controversies. For example, Chinese speakers are accustomed to use euphemisms to talk about topics such as death, sexual activities and disabilities, as explicit and direct narratives can be considered inappropriate or disrespectful.

In collecting the PETs, terms used by mainly ancient Chinese were excluded since the corpus is contemporary. Also, the PETs were restricted to single words and multi-word expressions, rather than sentences \cite{Zhang_2019}. The euphemistic terms are generated based on the language knowledge of the collector, who is a native speaker of Mandarin Chinese. For the source corpus, we referred to an online Chinese corpus made by Bright Xu (username: brightmart) on Github \cite{NLP_Chinese_Corpus}. The particular corpus used was \begin{CJK}{UTF8}{gbsn} 新闻语料json版 \end{CJK}(news2016zh) which consists of 2.5 million news articles from 63,000 media from 2014 to 2016, including title, keyword, summary and text body.

See Table \ref{tab:chinese} for examples of Chinese PETs. For example, \begin{CJK}{UTF8}{gbsn} 方便\end{CJK} means "to use the bathroom / to relieve oneself" when used euphemistically; and means "convenient" when used non-euphemistically. 

\subsubsection {Spanish}

%The history of the Spanish is conceived as an account of the ‘internal’ development of the language, a discussion of the way in which its phonology, its morphosyntax, its vocabulary and the meaning of its words have evolved and of the reasons for these developments(insofar as they can be established)~\cite{Penny-2022-Ralph}. 
Spanish, a Romance language, is the second most spoken language in the world \cite{ethnologue}. For the sake of building a wide and robust corpus, it was paramount considering all different dialects of Spanish. Some of the countries considered are: Equatorial New Guinea, Puerto Rico, Argentina, Spain, Chile, Cuba, Mexico, Bolivia, Ecuador, Paraguay, Dominican Republic, Venezuela, Costa Rica, Colombia, Nicaragua, Honduras, Guatemala, Per\'{u}, El Salvador, Uruguay, and Panama. 

Euphemisms are highly used in Spanish on a daily basis. Topics related to politics, employment, sexual activities or even death are widely communicated with euphemistic terms. First, a list of potentially euphemistic terms (PETs) was created using a dictionary of euphemisms as main reference \cite{lechado2000dicionario_eufemismos,rodriguez1999creatividad}. For extracting PETs, we relied heavily on the Real Academia Espa\~{n}ola (Real Spanish Academy)\footnote{\url{https://apps2.rae.es/CORPES/view/inicioExterno.view}}. The corpus we collected contains sentences with PETS, PET label (euphemistic/non-euphemistic), data source and country of origin. For example: "Pasar un buen rato" meaning "to have/spend a good time" can be used as both, euphemistically and non-euphemistically. This term could be used to express involvement on a sexual activity or to spend a good time with a friend, family or an acquainted. Furthermore, the phrase "Dar a luz" meaning "to give birth" is another example that comprises both uses. Women naturally give birth to babies but women can also give birth to wonderful ideas, so as any other human being. See more examples in Table \ref{tab:spanish}.

\begin{table*}[!h]
\begin{center}
\begin{tabularx}{\textwidth}{|P{3cm}|Y|Y|Y|Y|Y|Y|} 
 \hline
 \textbf{Language} & \textbf{Total Examples} & \textbf{Euph Examples} & \textbf{Non-Euph Examples} & \textbf{Total PETs} & \textbf{Always-Euph PETs} & \textbf{Ambiguous PETs}\\
 % \textbf{DRs} & \multicolumn{2}{c|}{\textbf{Formants}} & \multicolumn{2}{c|}{\textbf{MFCCs}} & \multicolumn{2}{c|}{\textbf{wav2vec2}}\\
 % \cline{2-7}
 % & ACC & F1 & ACC & F1 & ACC & F1 \\
 \hline
 American English & 1952 & 1383 & 569 & 129 & 71 & 58 \\
 \hline
 Mandarin Chinese & 1552 & 1134 & 418 & 70 & 46 & 24 \\
 \hline
 Spanish & 961 & 564 & 397 & 80 & 33 & 47 \\
 \hline
 {Yor\`{u}b\'{a}} & 1942 & 1281 & 661 & 129 & 62 & 69 \\
 \hline
\end{tabularx}
\end{center}
\caption{Statistics of multilingual datasets used for euphemism disambiguation experiments.}
\label{tbl:multilingual_stats}
\end{table*}

\begin{table*}[!h]
\begin{center}
\begin{tabularx}{\textwidth}{|P{3cm}|Y|Y|Y|Y|Y|Y|Y|Y|Y|} 
 \hline
 \textbf{Language} & \multicolumn{3}{c|}{\textbf{mBERT}} & \multicolumn{3}{c|}{\textbf{ XLM-RoBERTa-base}} & \multicolumn{3}{c|}{\textbf{ XLM-RoBERTa-large}} \\
 \cline{2-10}
 & \textbf{F1} & \textbf{P} & \textbf{R} & \textbf{F1} & \textbf{P} & \textbf{R} & \textbf{F1} & \textbf{P} & \textbf{R}\\
 \hline
 American English & 0.819 & 0.876 & 0.933 & 0.765 & 0.852 & 0.894 & 0.854 & 0.907 & 0.930\\
 \hline
 Mandarin Chinese & 0.901 & 0.952 & 0.938 & 0.884 & 0.921 & 0.960 & 0.952 & 0.967 & 0.982\\
 \hline
  Spanish & 0.747 & 0.781 & 0.816 & 0.765 & 0.799 & 0.819 & 0.776 & 0.813 & 0.826\\
 \hline
  {Yor\`{u}b\'{a}} & 0.729 & 0.801 & 0.859 & 0.683 & 0.771 & 0.843 & 0.667 & 0.768 & 0.814 \\
 \hline
\end{tabularx}
\end{center}
\caption{Results of euphemism disambiguation experiments on the multilingual datasets.}
\label{tbl:multilingual_results}
\end{table*}

\subsubsection{Yor\`{u}b\'{a}}

{Yor\`{u}b\'{a}} is one of the major languages of Nigeria, the most populous country on the African continent~\cite{Okanlawon2016ANAO}. With over 50 million language users as speakers, it is the third most spoken language in Africa~\cite{shode2022yosm}. There are many different dialects of Yoruba spoken by Yoruba people in Nigeria, Benin, and Togo, all of which are tonal (change depending on tone) and agglutinative (words are made up of linearly sequential morphemes) in nature.

Euphemisms are often used in everyday {Yor\`{u}b\'{a}} language conversations. Speakers use them to communicate sensitive topics like death and physical or mental health in a more socially acceptable manner, and to show reverence for certain people or occupations such as elders of the night which refer to witches and wizards, prostitutes, and so on. Euphemisms in {Yor\`{u}b\'{a}} are used to soften the harshness of situations; to report the death of an individual, speakers of the language mostly use indirect or subtle sentences instead of saying it directly. 

In NLP research, {Yor\`{u}b\'{a}} is considered as a low resourced language because of the limited availability of data in digital formats. There is no corpus dedicated to {Yor\`{u}b\'{a}} euphemisms available online so PETs were collected from different sources such as news websites like BBC {Yor\`{u}b\'{a}}, Alaroye, religious sources including {Yor\`{u}b\'{a}} Bible, \url{JW.org}, transcribed Muslim and Christian sermons, {Yor\`{u}b\'{a}} wikipedia, {Yor\`{u}b\'{a}} Web corpus (YorubaWaC), blogposts, journals, research works, books, Global Voices, Nigerian song lyrics, written texts written by {Yor\`{u}b\'{a}} native speakers 
and social media platforms such as tweets, Facebook public posts, and Nairaland. Some samples of PETs are listed in Table \ref{tab:yoruba}. 

\subsection{Methodology}

From each language dataset, a maximum of 40 euphemistic and non-euphemistic examples per PET were randomly chosen to be in the experimental dataset. This was done to in an effort to ensure an overall balance of PETs in the data and reduce skewed label proportions for each PET. We also include American English data, sampled in the same manner, to provide a basis of comparison. The final statistics for each dataset are shown in Table \ref{tbl:multilingual_stats}.

 We test three multilingual transformer models: mBERT \cite{devlin2018bert},  XLM-RoBERTa and  XLM-RoBERTa-large \cite{conneau2020unsupervised}. The hyperparameters used were the same as those described in \ref{section:vague_method}. A stratified 5-fold split is used to create 5 different train-test splits of each dataset, which includes every example while preserving the 80-20 ratio used in previous experiments.

\subsection{Results and Observations}

Table \ref{tbl:multilingual_results} shows the performance of each model. The metrics reported are macro-F1 (F1), precision (P), and recall (R), averaged across 5 experiments.

We note several things about the results: (1) All languages performed at least decently, indicating that multilingual BERT models pick up on something to disambiguate euphemisms in each language. (2) As expected,  XLM-RoBERTa-large generally performed better than  XLM-RoBERTa-base, which consistently performed worse than mBERT. (3) Because of differences in each language's dataset, the results are not directly comparable. We aim to make the experimental setup more consistent for future work, but some present inconsistencies include:
\begin{itemize}
    \item The Chinese data is the only one in which the PET is consistently "identified" (i.e. surrounded) by angle brackets <>, which the classifier may have used to its advantage. (Empirically, we notice that such "identifiers" improve performance.) 
    \item The proportion of non-euphemistic examples to the entire dataset was the smallest for Chinese (27\%), followed by English (29\%), {Yor\`{u}b\'{a}} (34\%) and Spanish (41\%). This, along with the number of ambiguous PETs, may reflect the relative "difficulty" of disambiguation for each language.
    \item While mBERT is pretrained on {Yor\`{u}b\'{a}} data, the  XLM-RoBERTa models are not. Thus, any sort of disambiguation capabilities shown by the  XLM-RoBERTa models are notable.
\end{itemize}

%\subsection{Appendices}
%Use \verb|\appendix| before any appendix section to switch the section numbering over to letters. See Appendix~\ref{sec:appendix} for an example.

\section{Conclusion and Future Work}

This study presents an expansion of the euphemism disambiguation task. We describe our method for annotating vagueness, and show that this kind of pragmatic distinction may reveal interesting trends in BERT's ability to perform NLU. Namely, BERT performs better for PETs labeled as VETs, which leads us to the potential result that BERT may be associating the presence of "sensitive words" to euphemisms. Corroborating this result and exploring additional properties of euphemisms are left for future work. 

The multilingual results show that BERT models can already disambiguate euphemisms in multiple languages to some extent, and establish a baseline from which to improve results. While continuously expanding the multilingual corpora is a must, a number of modeling aspects can be investigated as well. For instance, error analyses can be run to reveal potential misclassification trends in each language, and data and modeling improvements that were shown to work for American English can be attempted on other languages. In general, such investigations may be used to suggest useful cross-lingual features for PET disambiguation, and more broadly, universal properties of euphemisms.

\section*{Limitations}
Euphemisms are culture and dialect-specific, and we do not necessarily investigate the full range of euphemistic terms and topics covered by our selected languages. Even for "English", for instance, we do not explore euphemisms unique to "British English", though that warrants a study of its own. Additionally, as aforementioned, differences in the multilingual dataset render the results not directly comparable. For example, there are few large, structured corpora of {Yor\`{u}b\'{a}}, so the data was taken from a variety of sources, as opposed to the other languages. Additional limitations prevent some analyses, such as limited ability to identify the PET in {Yor\`{u}b\'{a}} due to loss of diacritics.

%ACL 2023 requires all submissions to have a section titled ``Limitations'', for discussing the limitations of the paper as a complement to the discussion of strengths in the main text. This section should occur after the conclusion, but before the references. It will not count towards the page limit.
%The discussion of limitations is mandatory. Papers without a limitation section will be desk-rejected without review.

%While we are open to different types of limitations, just mentioning that a set of results have been shown for English only probably does not reflect what we expect. 
%Mentioning that the method works mostly for languages with limited morphology, like English, is a much better alternative.
%In addition, limitations such as low scalability to long text, the requirement of large GPU resources, or other things that inspire crucial further investigation are welcome.

\section*{Ethics Statement}
The authors foresee no ethical concerns with the
work presented in this paper.
%Scientific work published at ACL 2023 must comply with the ACL Ethics Policy.\footnote{\url{https://www.aclweb.org/portal/content/acl-code-ethics}} We encourage all authors to include an explicit ethics statement on the broader impact of the work, or other ethical considerations after the conclusion but before the references. The ethics statement will not count toward the page limit (8 pages for long, 4 pages for short papers).

\section*{Acknowledgements} %only if accepted
This material is based upon work supported by the National Science Foundation under Grant  numbers: 2226006 and 1704113.

\bibliographystyle{acl_natbib}
\bibliography{anthology,custom,euphemisms,euphs}

\newpage
\onecolumn
\appendix
\section{List of Words used to Represent Sensitive Topics}
\bigskip
Listed below are the 22 "sensitive words" used to compute a sensitivity score for each example in the corpus:

\bigskip
\noindent
\label{sec:appendix_a}
['politics', 'death', 'kill', 'crime', 'drugs', 'alcohol', 'fat', 'old', 'poor', 'cheap', 'sex', 'sexual', 'employment', 'job', 'disability', 'pregnant', 'bathroom', 'sickness', 'race', 'racial', 'religion', 'government']

%\section{Example Appendix}
%\label{sec:appendix}

%This is a section in the appendix.

\end{document}